\documentclass[10pt,twocolumn,letterpaper]{article}

\usepackage{iccv}
\usepackage{times}
\usepackage{epsfig}
\usepackage{graphicx}
\usepackage{amsmath}
\usepackage{amssymb}
\usepackage{multirow}

\usepackage[pagebackref=true,breaklinks=true,letterpaper=true,colorlinks,bookmarks=false]{hyperref}
\usepackage{array}
\newcommand{\PreserveBackslash}[1]{\let\temp=\\#1\let\\=\temp}
\newcolumntype{C}[1]{>{\PreserveBackslash\centering}p{#1}}
\newcolumntype{R}[1]{>{\PreserveBackslash\raggedleft}p{#1}}
\newcolumntype{L}[1]{>{\PreserveBackslash\raggedright}p{#1}}

\iccvfinalcopy

\def\iccvPaperID{random}

\ificcvfinal\fi

\begin{document}

\title{PANet: Perspective-Aware Network with Dynamic Receptive Fields and Self-Distilling Supervision for Crowd Counting}

\author{Xiaoshuang Chen\\
{\tt\small chenxiaoshuang@sjtu.edu.cn}
\and
Yiru Zhao\\
{\tt\small yiru.zyr@alibaba-inc.com}
\and
Yu Qin\\
{\tt\small dongdong.qy@alibaba-inc.com>}
\and
Fei Jiang\\
{\tt\small jiangf@sjtu.edu.cn}
\and
Mingyuan Tao\\
{\tt\small juchen.tmy@alibaba-inc.com}
\and
Xiansheng Hua\\
{\tt\small xiansheng.hxs@alibaba-inc.com}
\and
Hongtao Lu\\
{\tt\small htlu@sjtu.edu.cn}
}

\maketitle

\ificcvfinal\thispagestyle{empty}\fi

\begin{abstract}
Crowd counting aims to learn the crowd density distributions and estimate the number of objects (e.g. persons) in images. The perspective effect, which significantly influences the distribution of data points, plays an important role in crowd counting. In this paper, we propose a novel perspective-aware approach called PANet to address the perspective problem. Based on the observation that the size of the objects varies greatly in one image due to the perspective effect, we propose the dynamic receptive fields (DRF) framework. The framework is able to adjust the receptive field by the dilated convolution parameters according to the input image, which helps the model to extract more discriminative features for each local region. Different from most previous works which use Gaussian kernels to generate the density map as the supervised information, we propose the self-distilling supervision (SDS) training method. The ground-truth density maps are refined from the first training stage and the perspective information is distilled to the model in the second stage. The experimental results on ShanghaiTech Part\_A and Part\_B, UCF\_QNRF, and UCF\_CC\_50 datasets demonstrate that our proposed PANet outperforms the state-of-the-art methods by a large margin.
\end{abstract}

\section{Introduction}

Crowd counting is an important topic in computer vision and has been widely applied in many tasks, such as video surveillance, urban traffic management, and passenger flow volume statistics. 
CNN-based density estimation methods have achieved remarkable performance. 
However, there are still some challenging problems in crowd counting, such as the perspective effect.
Earlier works~\cite{conf/cvpr/ZhangZCGM16, conf/eccv/Onoro-RubioL16, conf/mm/BoominathanKB16, conf/cvpr/SamSB17, conf/iccv/SindagiP17, conf/ijcai/LiuWLOL18, conf/cvpr/0011GMH18} adopt multi-column networks to implement the variety of the perspective response in an image.
However, the receptive fields are still fixed for all images.
Recently proposed methods \cite{conf/mm/GuoLZW19, conf/cvpr/LiuLZNPW19} implement self-adaptive receptive fields with deformable convolution~\cite{conf/iccv/DaiQXLZHW17}.
\cite{conf/cvpr/BaiHQHWY20} corrects the annotations by EM algorithm and proposes adaptive dilated convolution. 
The dynamic receptive fields in these works are trained by end-to-end schemas thus are not explainable. 
In this paper, we propose a perspective-aware network called PANet, which addresses the perspective problem from the following two aspects. 

\begin{figure}[tp]
    \centering
    \includegraphics[width=1.0\linewidth]{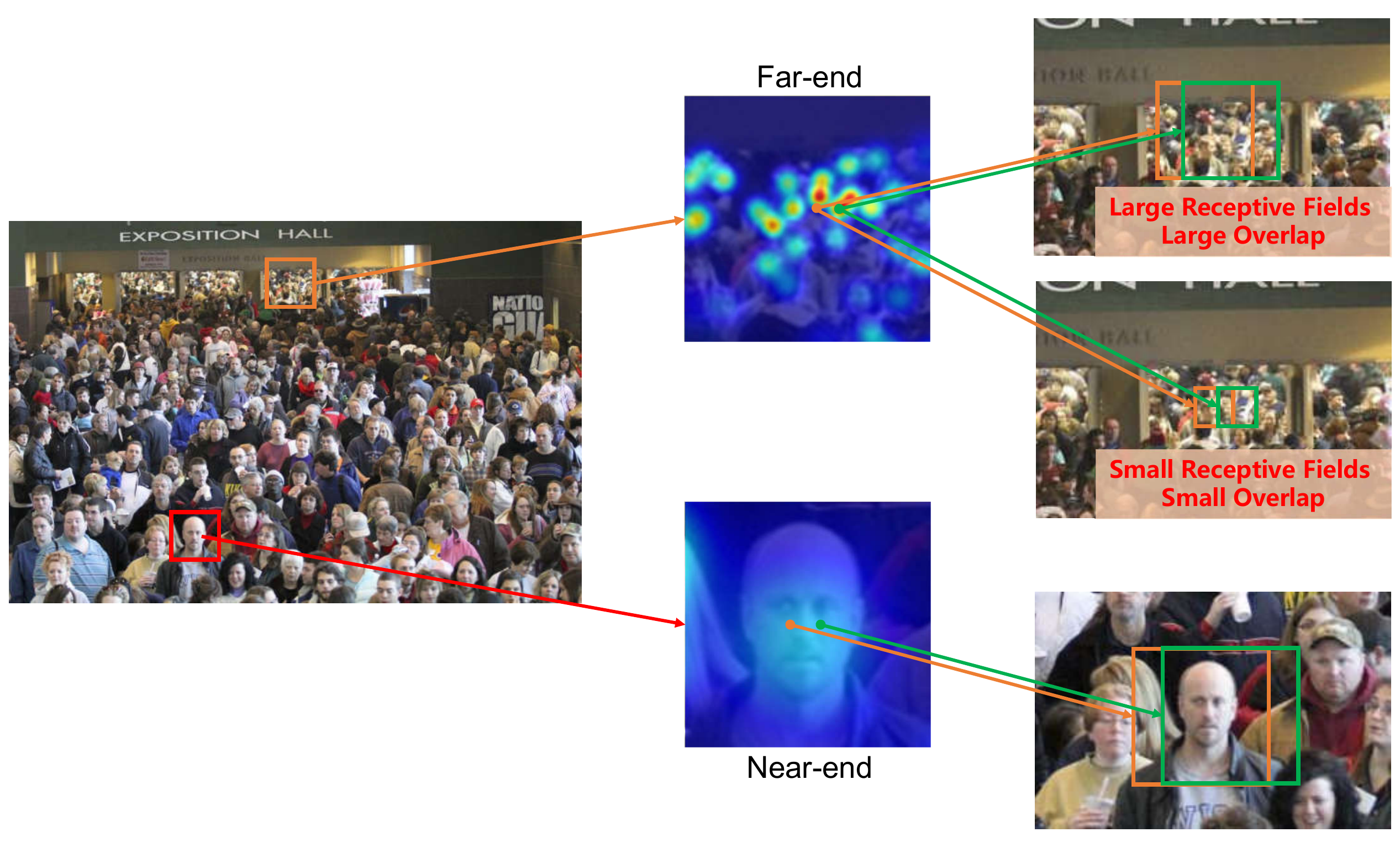}
    \caption{The illustration of the receptive fields with different scales of the perspective response. In far-end regions, the density variation is larger, and the overlap of the receptive fields of adjacent positions should be smaller for better discrimination. The density is smooth in near-end regions, and large receptive fields are adopted for enough context.}
    \label{fig:first}
\end{figure}

Firstly, we propose the dynamic receptive fields (DRF) learning framework, which creates an association between the scale of the receptive fields and the density distributions.
The adjustment of receptive field scales in crowd counting models is important to cope with the scale variation of different perspective areas. 
As illustrated in Fig.~\ref{fig:first}, the near-end objects are sparser than the far-end ones in the image due to the perspective effect.
In the commonly used fully-convolutional architecture, the network learns to regress the counting score for each point in the down-sampled heatmap based on the image information within the receptive field of this point. 
Considering two adjacent points in the output heatmap, the overlap of the receptive fields of these two points becomes larger as the receptive field increases.
In other words, larger receptive fields provide more similar input visual signals to the network for adjacent predicted points.
It can be observed that the supervised density map is smooth in the near-end area, but has obvious variations in the far-end area.
In the far-end area, the ground-truth may change dramatically in a local region. 
It is difficult for the network to estimate the variation of the counting score based on the highly overlapped input patches.
Thus, the model ought to apply small receptive fields in the far-end to learn more discriminative features for adjacent regions.
Meanwhile, large receptive fields ought to be used in the near-end to involve the necessary context of big objects. 
Based on this assumption, we adjust the receptive fields according to the density variation. 
A rough network is trained for learning a rough density estimation, which is transformed to the dilation map embedded in the precise network, i.e., the network for the count prediction, by a linear transformer. 

Secondly, we make improvements in perspective-aware supervision strategies. Commonly used crowd counting datasets provide point annotations as ground truth labels. Most previous state-of-the-art works~\cite{conf/cvpr/ZhangZCGM16, conf/cvpr/SamSB17, conf/cvpr/LiZC18, conf/eccv/CaoWZS18, conf/cvpr/SamSBS18,  conf/iccv/ZhangSXZZ0019, conf/cvpr/LiuLZNPW19} use Gaussian kernels to smooth annotated dots. 
However, applying Gaussian distributions is sub-optimal due to the constrained scale of response areas, especially when there is large scale variation due to the perspective effect. 
In this paper, we propose a two-stage training framework called self-distilling supervision (SDS). Motivated by knowledge distillation~\cite{conf/kdd/BucilaCN06, hinton2015distilling}, in the first learning stage, the precise network distills the perspective information and generates refined density maps that are more network-friendly and reflect the actual scale variation of the perspective response better. 
In the second stage, we train the same precise network with the refined density maps. The perspective information is distilled from the teacher network to the student network. 
SDS benefits the training process and achieves better evaluation performance than Gaussian-based supervision.

The main contributions of this paper are summarized as follows:

1. We propose a perspective-aware dynamic receptive fields (DRF) learning framework guided by rough density distributions, which is able to dynamically adjust the receptive field according to the input image. 
    
2. We propose a two-stage supervision framework called self-distilling supervision (SDS). The refined density map serves as a better supervision for the model training.
    
3. Comprehensive experiments on four datasets demonstrate that our approach outperforms state-of-the-art methods, e.g., 45.2 MAE on ShanghaiTech Part\_A, 5.9 MAE on ShanghaiTech Part\_B, 49.1 MAE on UCF\_QNRF and 160.3 MAE on UCF\_CC\_50.

\section{Related Work}

Early works such as \cite{conf/avss/TopkayaEP14, conf/icpr/LiZHT08, conf/cvpr/LeibeSS05, journals/pami/EnzweilerG09, conf/cvpr/BrostowC06, conf/cvpr/ZhaoN03, conf/cvpr/GeC09, conf/eccv/DollarBBPT08, journals/tsmc/LinCC01} consider crowd counting as a detection based problem. However, these approaches have poor estimation results on images with extremely dense crowds. In recent years, CNN-based density estimation methods~\cite{journals/eaai/FuXLLYZ15, conf/mm/WangZYLC15, conf/eccv/WalachW16} outperform the traditional detection based methods and have achieved considerable progress in dense crowd counting. In this section, we mainly review two mainstream improvement aspects for crowd counting.

\subsection{Network Architectures}
Some works improve the performance by designing better network architectures to adjust the receptive fields of the models. 

MCNN~\cite{conf/cvpr/ZhangZCGM16} utilizes three branches with multi-size convolution kernels, aiming at solving the scale variation. Hydra-CNN~\cite{conf/eccv/Onoro-RubioL16} learns a multi-scale model which uses a pyramid of image patches with different scales. CrowdNet~\cite{conf/mm/BoominathanKB16} combines shallow and deep network branches, capturing the high-level and the low-level features. In Switch-CNN~\cite{conf/cvpr/SamSB17}, a switch layer is designed to select the best regression network. In CP-CNN~\cite{conf/iccv/SindagiP17}, the model incorporates global and local context. Adversarial loss and pixel-level loss are combined for fusing all the features. DRSAN~\cite{conf/ijcai/LiuWLOL18} proposes a Spatial Transformer Network to cope with variations in scales and rotations. DecideNet~\cite{conf/cvpr/0011GMH18} generates the detection-based and regression-based density maps respectively. TDF-CNN~\cite{conf/aaai/SamB18} proposes a bottom-up network along with a separate top-down network to generate feedback.  McML~\cite{conf/mm/ChengLD0HH19} estimates mutual information between columns. DADNet~\cite{conf/mm/GuoLZW19} takes advantage of dilated-CNN and adaptive deformable convolution. D-ConvNet~\cite{conf/cvpr/ShiZLCYCZ18} proposes a new learning strategy called deep negative correlation learning. CSRNet~\cite{conf/cvpr/LiZC18} employs dilated convolution to expand the receptive fields. In SANet~\cite{conf/eccv/CaoWZS18}, the encoder extracts features with different receptive fields using scale aggregation modules, and transposed convolution is used in the decoder. ADCrowdNet~\cite{conf/cvpr/LiuLZNPW19} combines an attention map generator and multi-scale deformable convolutional scheme. TEDNet~\cite{conf/cvpr/JiangXZZ0D019} is a trellis encoder-decoder network that contains multiple decoding paths to aggregate features. CAN~\cite{conf/cvpr/LiuSF19} encodes multi-level contextual information into an end-to-end pipeline adaptively. S-DCNet~\cite{conf/iccv/XiongLLLCS19} divides dense regions sub-region counts are within a closed set. RPNet~\cite{conf/cvpr/YangLWSHS20} proposes a reverse perspective network to correct the perspective distortions.

\begin{figure*}[tp]
    \centering
    \includegraphics[width=0.9\linewidth]{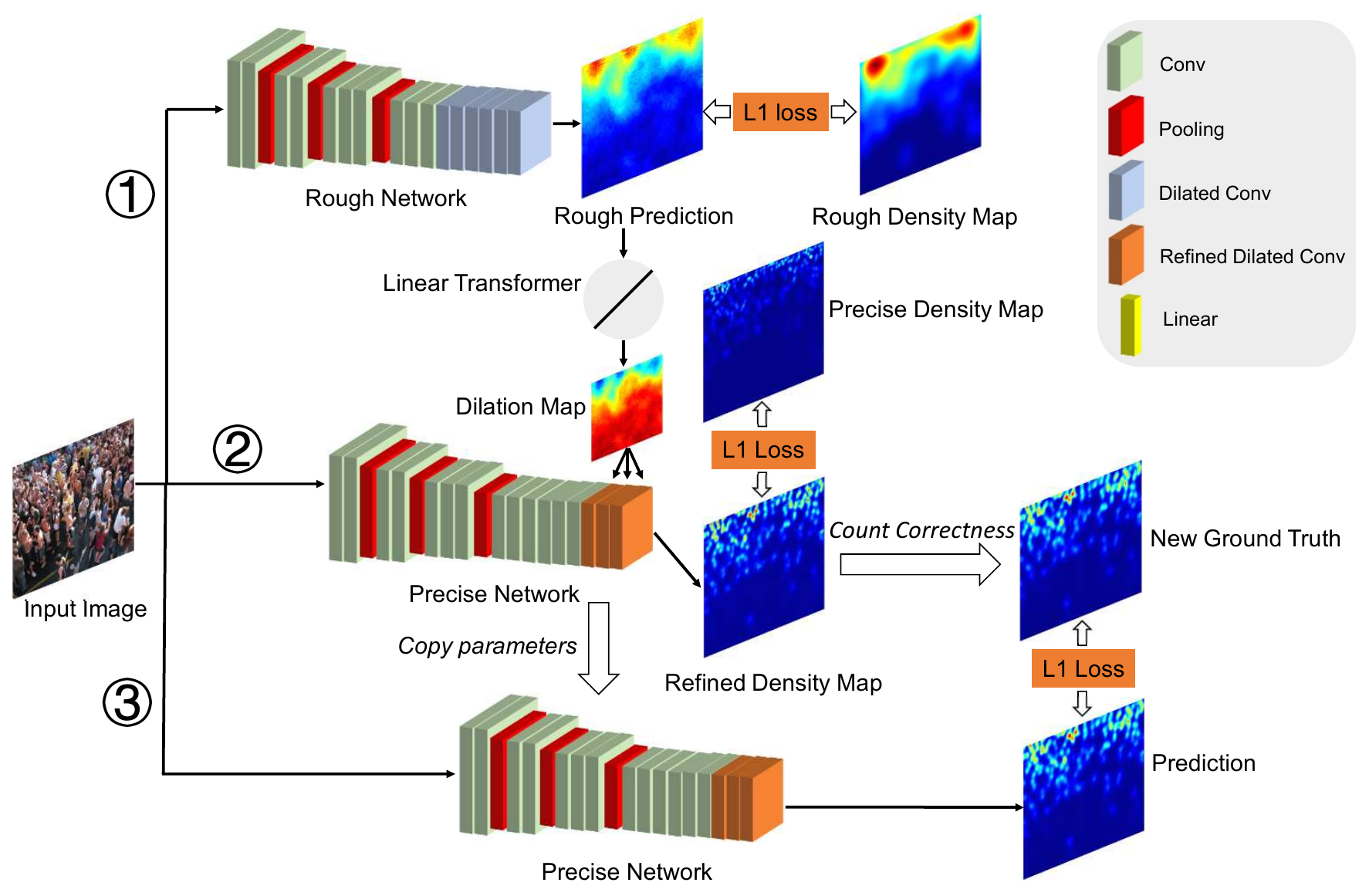}
    \caption{Overview of PANet. The training process can be divided into three steps. (1) The input image is fed into the rough network to generate a rough prediction first, supervised by the rough density map. After a linear transformer, the rough prediction is used as a dilation map in the refined dilated convolution layers of the precise network. (2) The learning of the precise network can be divided into two stages. Firstly the input image is fed into the precise network to distill the perspective information by upgrading the precise density map. (3) After count correctness, the refined density map with the perspective information distilled from the first stage serves as the new ground truth for training the precise network. The two training stages of the precise network share the same network structure.}
    \label{fig:model}
\end{figure*}

However, the prior knowledge that the scale of receptive fields is corresponding to the perspective information is not well utilized yet.
Based on this assumption, our dynamic receptive fields framework learns to adjust the receptive fields according to the density variation.

\subsection{Supervision Strategies}

Recently, methods based on effective supervision strategy have achieved significant progress in dense crowd counting. IG-CNN~\cite{conf/cvpr/SamSBS18} introduces a growing CNN tree to adapt to the wide variability seen. ACSCP~\cite{conf/cvpr/ShenXNWHY18} employs an adversarial loss and a scale-consistency regularizer. CFF~\cite{conf/iccv/ShiMS19} obtains a focus from segmentation and a focus from global density besides density maps from point annotations. In RAZ-Net~\cite{conf/cvpr/LiuWM19}, a recurrent attentive zooming network is proposed for precise localization. There are some works which apply different loss functions to optimize models better, such as \cite{conf/cvpr/ShenXNWHY18, conf/iccv/ChengLD0H19, conf/cvpr/JiangZXZLZYP20, journals/corr/abs-2009-13077, conf/iccv/MaWHG19}. ACSCP~\cite{conf/cvpr/ShenXNWHY18} employs an adversarial loss and a scale-consistency regularizer. In SPANet~\cite{conf/iccv/ChengLD0H19}, Maximum Excess over Pixels loss is proposed to incorporate spatial context. Other methods are exploring potential supervision information or generating new density maps as labels from point annotations, such as ~\cite{conf/iccv/ShiMS19, conf/cvpr/BaiHQHWY20, wan2020modeling, conf/cvpr/LiuSZW19}. CFF~\cite{conf/iccv/ShiMS19} obtains a focus from segmentation and a focus from global density besides density maps from point annotations. 

However, the manual-designed density maps are sub-optimal for training the model.
In this paper, we propose self-distillation supervision and achieve remarkable progress by the perspective information distilled from the teacher model to the student model.

\section{Methodology}
\label{sec:method}

We propose a perspective-aware framework for crowd counting named PANet, which is illustrated in Fig.~\ref{fig:model}. In this Section, we introduce the two components of our PANet respectively. First, we provide the principle of our dynamic receptive fields (DRF) framework and then we present our self-distilling supervision (SDS) framework.

\subsection{Dynamic Receptive Fields}
\label{sec:rrf}

To address the scale variation of the receptive fields in scenes with large density variations in crowd counting, we propose a perspective-aware learning framework called dynamic receptive fields (DRF). 
Due to the perspective effect, the objects in the near-end regions are sparse than those in the far-end regions.
The overlap of the receptive fields of adjacent positions becomes larger with the scale of the receptive fields, as illustrated in Fig.~\ref{fig:first}. We denote the overlap of the receptive fields by the Intersection over Union (IoU) to quantify this observation.
Assume that the distance between two positions in one row of the output density map is $n$ pixels and the down-sample ratio of the precise network is $k$. The variation of the IoU about the scale of receptive fields in the precise network is formulated as:
\begin{equation}
    \mathrm{IoU} = \frac{X-kn}{X+kn}, X\geq kn
\end{equation}
where $X$ is the scale of the receptive fields. Thus, the IoU of the receptive fields at a pair of fixed positions increases with the scale of the receptive fields. In more congested regions in an image, the variation of the density map is more dramatic, while the large overlap of the adjacent receptive fields restricts the model to discriminatively estimate the counting score. Thus, the IoU of the receptive fields needs to be smaller in the dense area.
So we propose the dynamic receptive fields, which are smaller in the denser regions and are larger in sparser regions to involve necessary context.
To validate the rules above, we implement some comparison experiments with different dilated convolution methods and different ground truth annotations in Sec.~\ref{sec:ablation}.

The proposed DRF adjusts the scale of receptive fields by the dilated convolution parameters corresponding to density. As illustrated in Fig.~\ref{fig:model}, the rough density map is generated using Gaussian kernels with large spreads. 
The rough network is trained to predict the rough density map, using $L_1$ distance as the loss function with the same definition in Eq.~\ref{eq:l1}. 
We use a linear transformer module to transform the rough prediction to the dilation map. Instead of a fixed dilation rate, the dilation map is embedded in the last three layers of the precise network, named refined dilated convolutional layers. Thus, the dilation rates are different in regions with different rough density estimations when training the precise network.

The dilation rates in the output dilation map are transformed to $[0, R]$ by a negative linear transformer, which means the receptive fields are smaller in the denser regions.
The operation of the linear transformer is formulated as:
\begin{equation}
    L(x) = \begin{cases}R, & x \leq 0\\
    R - \gamma x, & 0 < x < \frac{R}{\gamma}\\
    0, & x \geq \frac{R}{\gamma}\end{cases},
\end{equation}
where $R$ is the pre-set upper bound, $\gamma$ is a positive coefficient, and $x$ is the rough prediction of the rough network.

The value at the $k$th position in the output feature map of the refined dilated convolutional layer $F_{out}(x_k)$ is calculated as:
\begin{equation}
F_{out}(x_k) = \sum_{i, j \in D}F_{in}(x_k + \hat{r}\times (i, j))\times w_{i, j},
\end{equation}
where $\hat{r}$ is the dilation rate at the $k$th position computed by $L(x)$, $w$ denotes the weights in convolution kernels, and $D$ is the collection of unitized offsets defined as:
\begin{equation}
    D = \{-1, 0, 1\}.
\end{equation}
Bilinear interpolation is used to calculate the values at continuous sampling points. 

With the refined dilated convolutional layers, the precise network is able to adjust the receptive fields for different local regions based on the input image, and learns to predict more discriminative estimations.

\subsection{Self-Distilling Supervision}

The traditional learning strategies of Crowd Counting often use Gaussian density maps as the ground truth.
However, the Gaussian annotations have constrained scales of response areas. This drawback is particularly obvious when there is a large density variation in images. These density maps need to be refined for better learning of the perspective effect. Motivated by knowledge distillation, we propose a two-stage learning framework called self-distilling supervision (SDS), considering that the predicted density map is more aligned with the actual scales of the response areas and is more sensitive to the variation of the perspective information.

\begin{figure}[tp]
    \includegraphics[width=1.0\linewidth]{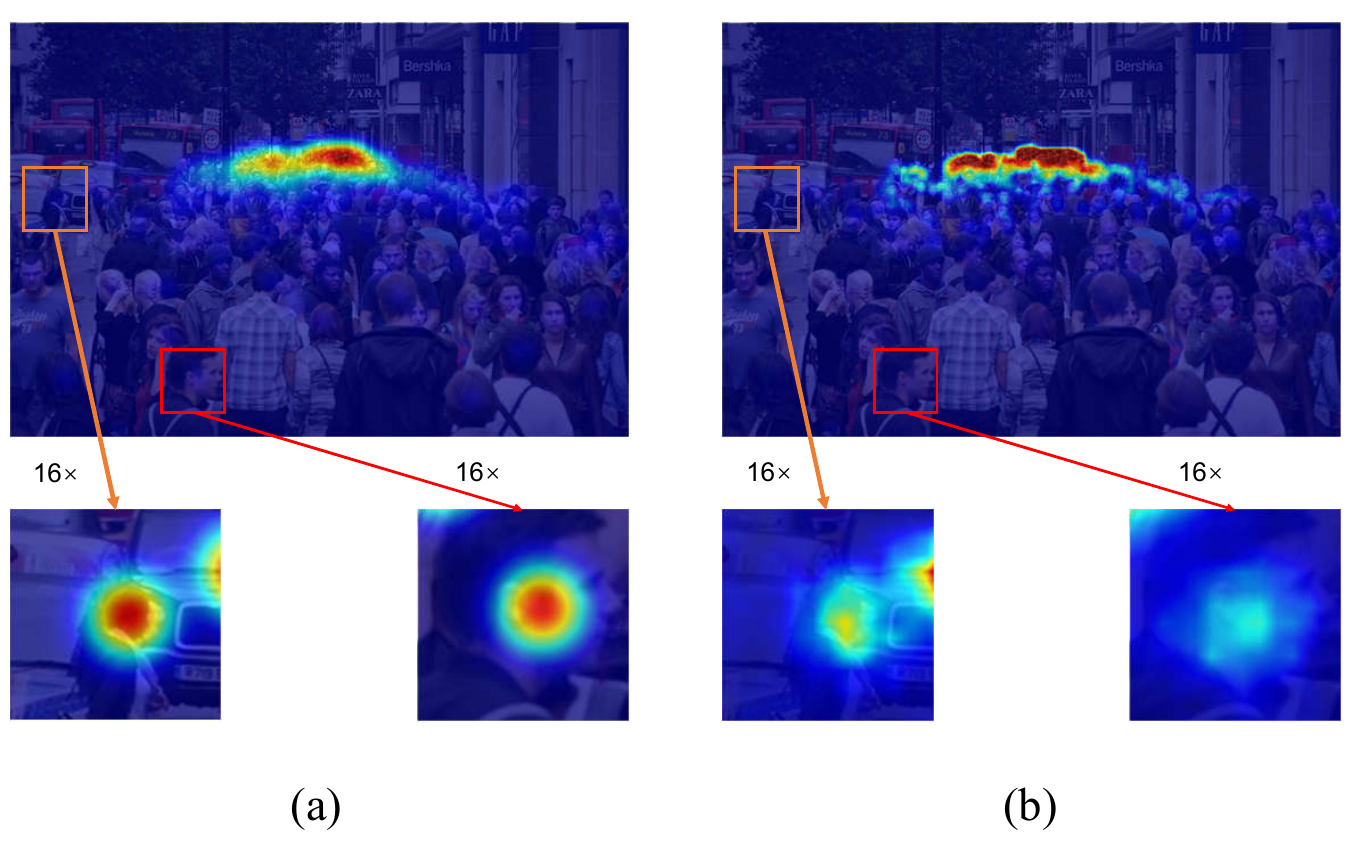}
    \caption{Visualization examples of different ground truth density maps. (a) The Gaussian annotated density maps. (b) Our refined density maps. The density maps are normalized in each local patch for better visualization.}
    \label{fig:rs}
\end{figure}

Our SDS is a two-stage learning framework, where the purpose of the first stage is to distill the perspective information extracted from the teacher network to the student network.
The teacher and student networks share the same architecture of the precise network, as shown in Fig.~\ref{fig:model}, so our method is called self-distilling. In the first stage, the teacher network is trained to estimate the crowd counting, supervised by the precise density maps, which are smoothed by Gaussian kernels with small spreads.
This stage is the same as other CNN-based density estimation methods.
A pixel-wise loss function is needed for training the network. 
The $L_1$ distance is adopted in our method, which is defined as:
\begin{equation}
\label{eq:l1}
    \mathcal{L}_{1}(\hat{Y}, Y) = \frac{1}{n} \sum_{i=1}^{n}\left|\hat{y}_i-y_i\right|.
\end{equation}
An interesting observation is that the refined density map predicted by the teacher network is self-adaptive and reflects the actual perspective information after training of deep CNN.
Fig.~\ref{fig:rs} shows visualization examples of both the Gaussian annotations and the refined density maps for comparison. The refined density maps have more reasonable density variations according to perspective information and the target objects.
Thus the refined density maps serve as better supervision targets than the Gaussian annotations and are utilized to optimize the student network in the second learning stage. 

Although the refined density map contains sufficient perspective information, it can not be used as the ground-truth heatmap directly due to the inevitable counting error caused by the teacher network.
Thus we need to correct the total count of the refined density map while keeping the relative data distribution by the count correctness process, which is formulated as:
\begin{equation}
    h_i = \frac{\sum_{j=1}^{n}y_j}{\sum_{j=1}^{n}\hat{y}_j}\hat{y}_i
\end{equation}
where $\hat{y}_i$ is the $i$-th value of the refined density map predicted by the teacher network, and $y_i$ is the $i$-th value of the Gaussian annotation heatmap.
Then the total count in the refined density map becomes identical to the original annotations. After this adjustment, the student network is trained with $L_1$ loss again with new supervision targets for the final count estimation. 

The refined density map facilitates the supervision process and leads to higher prediction accuracy because it utilizes self-distillation to upgrade the supervision targets. The perspective information is distilled in the first stage and has better guidance for the training of the model.

\section{Experiments}

\subsection{Datasets}

We validate the effectiveness of our PANet on four challenging crowd counting datasets, including ShanghaiTech~\cite{conf/cvpr/ZhangZCGM16} Part\_A and Part\_B, UCF\_QNRF~\cite{conf/eccv/IdreesTAZARS18} and UCF\_CC\_50~\cite{conf/cvpr/IdreesSSS13}.

\textbf{ShanghaiTech Part\_A}~\cite{conf/cvpr/ZhangZCGM16} contains 482 images randomly crawled from the Internet, in which 300 images and 182 images are used as the training and test sets respectively. There are congested scenes in most of the images. There is a wide range of density variations in images.

\textbf{ShanghaiTech Part\_B}~\cite{conf/cvpr/ZhangZCGM16} contains 716 images of sparse scenes, taken from busy streets in Shanghai. There are 400 images in the training set and 316 images in the test set. There is a small number of head annotations in most of the images.

\textbf{UCF\_QNRF}~\cite{conf/eccv/IdreesTAZARS18} contains 1535 images and is divided into training the training set with 1201 images and the test set with 334 images. This dataset has a large number of head annotations.

\textbf{UCF\_CC\_50}~\cite{conf/cvpr/IdreesSSS13} collects only 50 images and has a large span in object count among images, so it is an especially challenging dataset. 5-fold-cross-validation is used following \cite{conf/cvpr/IdreesSSS13}.

\subsection{Training Details}

\begin{table*}[tp]
    \centering
    \caption{Ablation study of DRF and SDS on four datasets.}
    \begin{tabular}{c|cc|cc|cc|cc}
        \hline
        \multirow{2}{*}{Methods} &\multicolumn{2}{c|}{ShanghaiTechA} &\multicolumn{2}{c|}{ShanghaiTechB}&\multicolumn{2}{c|}{UCF\_QNRF}&\multicolumn{2}{c}{UCF\_CC\_50}\\ 
        & MAE  & RMSE  &MAE &RMSE &MAE &RMSE &MAE &RMSE \\
        \hline
        Baseline & 57.4 & 88.9 & 7.6 & 12.1 & 115.9 & 185.4 &250.1&335.4\\
        SDS & 53.4 & 85.5 & 7.0 & 11.0 & 101.9 & 170.5 &192.6&269.1\\
        DRF & 47.1 & 74.3 & 6.2 & 10.1 & 79.6 & 135.6 &211.1&273.6\\
        PANet & \textbf{45.2} & \textbf{73.1} & \textbf{5.9} & \textbf{9.3} & \textbf{49.1} & \textbf{106.0} & \textbf{160.3} & \textbf{223.7}\\
        \hline
    \end{tabular}
    \label{tab:ablation}
\end{table*}

For datasets with high-density crowds such as ShanghaiTech Part\_A, UCF\_QNRF, UCF\_CC\_50 datasets, we augment the training data using
horizontal flipping and cropping with crop size $256\times 256$. The batch size is set to 32. For ShanghaiTech Part\_B, the crop size is set to $512\times 512$ and the batch size is set to 8 considering the low-density crowds in this dataset. For UCF\_QNRF, considering the large size of images, we resize the longer side to 1024 pixels with the constant aspect ratio. There is no data augmentation except resizing during the testing process.

Adam Optimizer~\cite{journals/corr/KingmaB14} is applied with a fixed learning rate at $10^{-5}$ and weight decay at $10^{-4}$ in the training process. In the second training stage of SDS, we copy the precise network parameters from the first stage and finetune it with the new supervision targets.

\begin{table}[tp]
    \centering
    \caption{Performance of DRF with different $\gamma$ in the linear transformer on ShanghaiTech Part\_A.}
    \begin{tabular}{c|ccccc}
    \hline
        $\gamma$ & 1 & 5 & 10 & 15 & 20 \\
        \hline
        MAE  & 56.2 & 48.9 & \textbf{47.1} & 49.4 & 52.0 \\
        RMSE  & 89.1 & 75.0 & \textbf{74.3} & 82.5 & 83.2 \\
        \hline
    \end{tabular}
    
    \label{tab:gamma}
\end{table}

\begin{table}[tp]
    \centering
    \caption{Performance of different methods with dilated convolution on ShanghaiTech Part\_A.}
    \begin{tabular}{c|cc}
    \hline
        Methods & MAE  & RMSE  \\
        \hline
        Dila. 1 & 59.2 & 96.5\\
        Dila. 2 & 57.4 & 88.9\\
        Deformable Conv.~\cite{conf/iccv/DaiQXLZHW17} & 59.4 & 92.9\\
        Adaptive Dila.~\cite{conf/cvpr/BaiHQHWY20} & 56.8& 94.5\\
        DRF & \textbf{47.1} & \textbf{74.3}\\
         \hline
    \end{tabular}
    
    \label{tab:dilation}
\end{table}

Our baseline is CSRNet~\cite{conf/cvpr/LiZC18} with first ten convolutional layers of VGG16\_bn~\cite{journals/corr/SimonyanZ14a, conf/icml/IoffeS15} pretrained on ImageNet~\cite{journals/ijcv/RussakovskyDSKS15}, labeled as CSRNet*, which is also our backbone of both the precise and the rough network. To guarantee the efficiency and the lightweight of PANet, only the last three layers of the precise network are replaced by our refined dilated convolutional layers. Even though our DRF is only applied to a small portion of the precise network, the estimation performance is improved remarkably.

The original precise ground truth density maps are generated by a fixed spread Gaussian kernel for ShanghaiTech Part\_B, UCF\_QNRF, and UCF\_CC\_50, and we set $\sigma=15$. We use the K-Nearest Neighbors (KNN) algorithm to compute the adaptive spread of Gaussian kernels for ShanghaiTech Part\_A following \cite{conf/cvpr/ZhangZCGM16}. To construct the rough density maps, we set $\sigma=50$ in the Gaussian kernel for all four datasets. The output density map is $1/8$ size of the original input image, so we downsample the ground truth density map while keeping the total count constant.

The Mean Absolute Error (MAE) and the Root Mean Squared Error (RMSE) are used as our evaluation metrics, following previous works, which are computed as follows:
\begin{equation}
    \mathrm{MAE} = \frac{1}{N}\sum_{i=1}^{N}\left|C'_i - C_i\right|, 
\end{equation}
\begin{equation}
    \mathrm{RMSE} = \sqrt{\frac{1}{N}\sum_{i=1}^{N}\left|C'_i - C_i\right|^2},
\end{equation}
where $C_i$ is the ground truth count and $C'_i$ is the predicted count for the $i$th image, and $N$ is the number of image samples.

In the test stage, the input image is fed into the rough network first to generate the dilation map that is embedded in the refined dilated convolutional layers. Then the precise network outputs the final estimation result.

\begin{figure}[tp]
    \centering
    \includegraphics[width=1.0\linewidth]{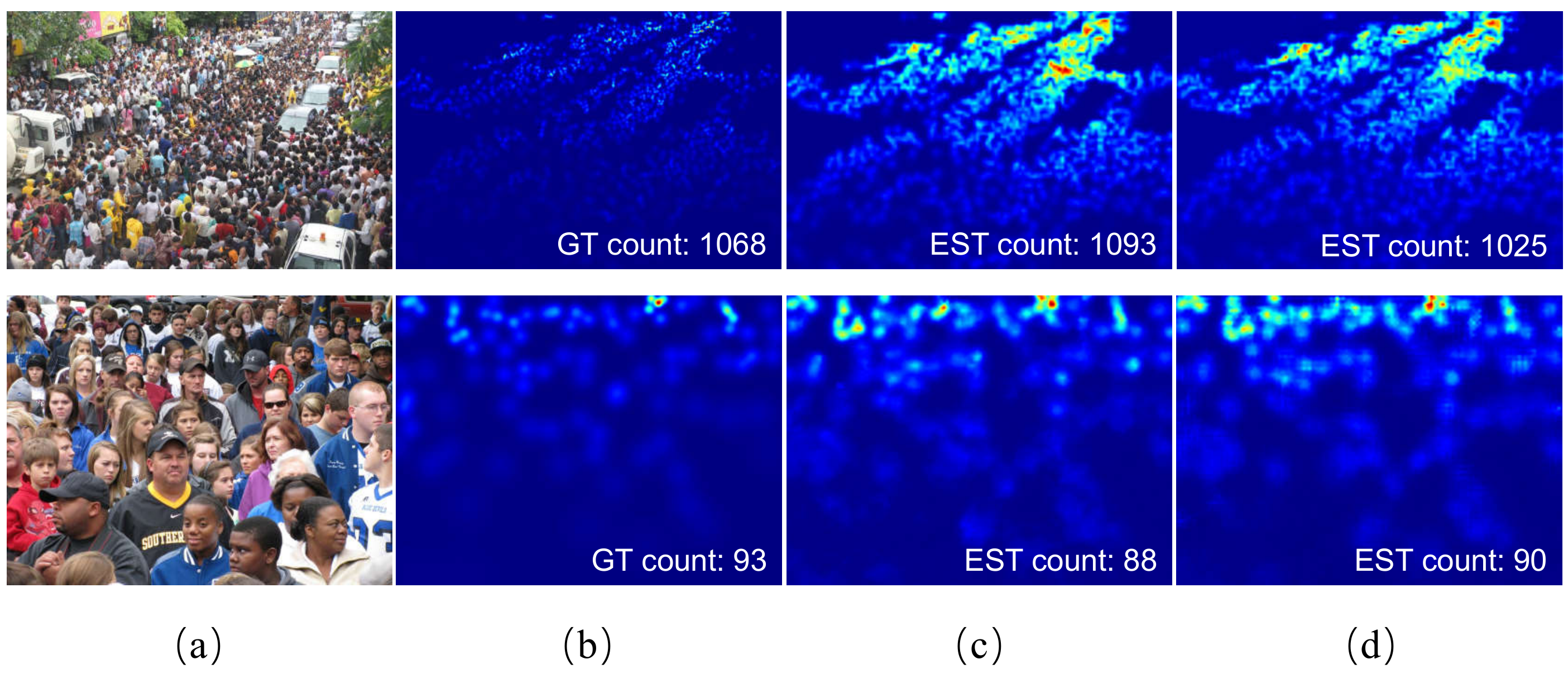}
    \caption{Visually qualitative results of different dilation rates on ShanghaiTech Part\_A. The four columns are: (a) input images, (b) Gaussian ground truth density maps, (c) the baseline with dilation rate as 1, (d) the baseline with dilation rate as 2. Our baseline with a smaller dilation rate has more accurate estimations on congested scenes than the baseline with a larger dilation rate, while the latter is superior to the former on sparse scenes.}
    \label{fig:dilate12}
\end{figure}

\begin{figure*}[tp]
    \centering
\includegraphics[width=0.7\linewidth]{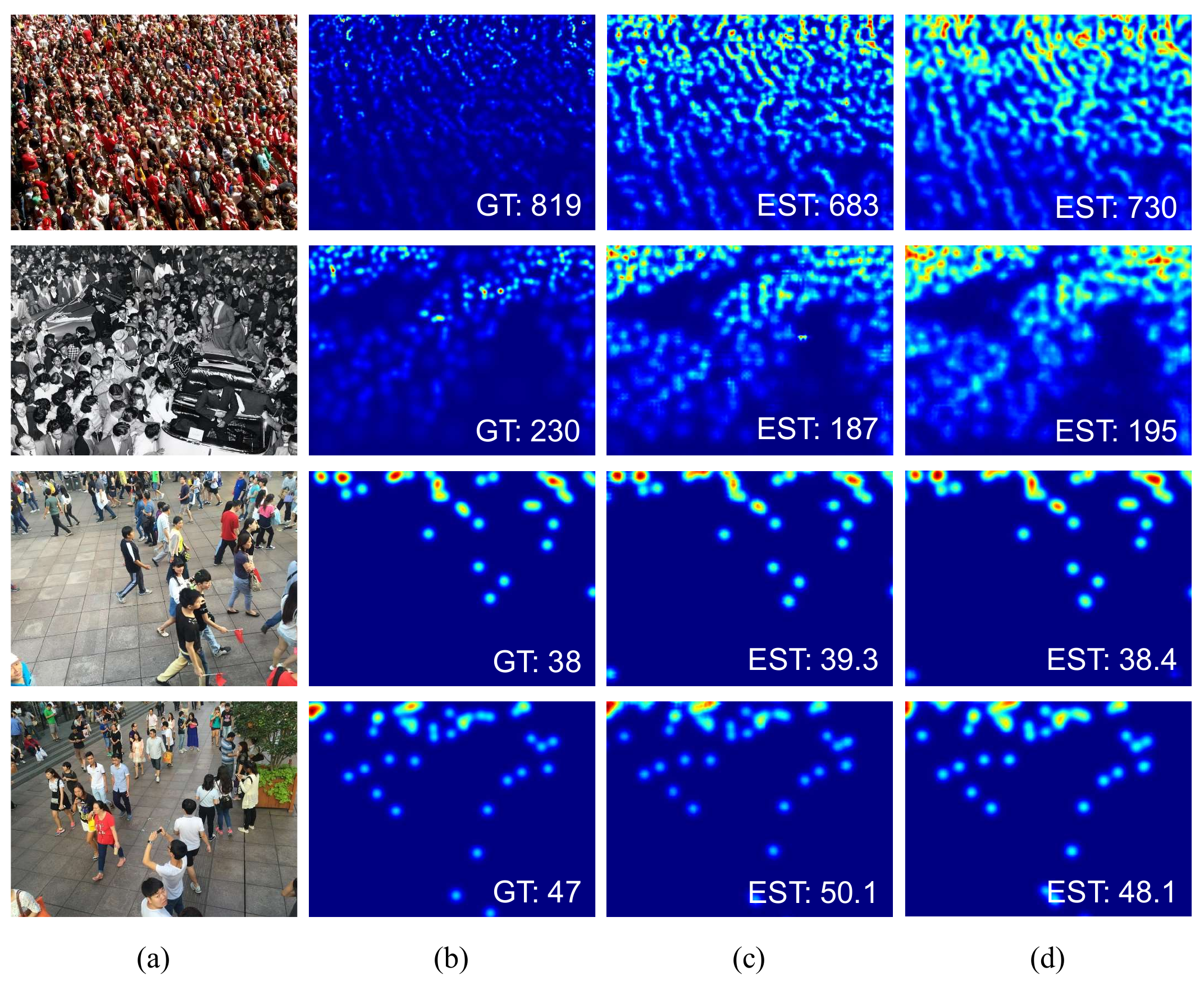}
    \caption{Visualization of estimated density maps on the ShanghaiTech dataset. From left to right, we display (a) input images, (b) Gaussian ground truth density maps, (c) estimations by our baseline CSRNet*, (d) estimations by PANet.}
    \label{fig:perform}
\end{figure*}

\subsection{Ablation Experiments}
\label{sec:ablation}

To validate the motivation of the DRF, i.e., the variation of the receptive scale according to the densities, we set different dilation rates in the baseline and visualize some predicted density maps in Fig.~\ref{fig:dilate12}. It is obvious that although the precise network with the dilation rate of 1 has a larger count estimation bias on sparse scenes due to the restricted receptive fields, it predicts densities in congested regions more precisely, compared with the one with the dilation rate of 2. There is a wide range of density variations in plenty of scenes, so a fixed dilation rate leads to sub-optimal performance. Therefore, reducing the receptive fields in far-end regions and enlarging them in near-end regions are reasonable assumptions.

We also test different $\gamma$ in the linear transformer. The comparison experiments on ShanghaiTech Part\_A are illustrated in Tab.~\ref{tab:gamma}. The MAE and RMSE decrease with the increase of $\gamma$ at first and increase at last. We choose the coefficient $\gamma=10$ for better performance of DRF.

To evaluate the effectiveness of DRF compared with other methods involving dilated convolution, we carry out ablation experiments on ShanghaiTech Part\_A, as shown in Tab.~\ref{tab:dilation}. The methods include dilated convolution with fixed dilation rate 1 and 2, the adaptive dilated convolution~\cite{conf/cvpr/BaiHQHWY20} and our DRF.
The deformable convolution~\cite{conf/iccv/DaiQXLZHW17} and the adaptive dilated convolution~\cite{conf/cvpr/BaiHQHWY20} learn the dilation maps implicitly from end-to-end training and are inexplicable. In our DRF learning framework, the dilation map is trained separately with the rough density map as the supervision targets considering the perspective effect. In consequence, DRF outperforms the other methods by large margins.

Our DRF improves the estimation accuracy by reducing the scale of the receptive fields in regions where the ground truth varies dramatically and vice versa, so the methods of annotations may have an influence on the performance of our DRF. There are two main Gaussian annotation methods, one is fixed $\sigma$ and the other is adaptive $\sigma$. The adaptive $\sigma$ annotation method, e.g., constructing kernels by KNN algorithm, has the more obvious perspective response, i.e., there is a significant difference in the amplitude of ground truth variations between regions with different densities. The experiment results are shown in Tab.~\ref{tab:annotation}. It can be observed that our proposed DRF brings larger improvements on baseline when using the adaptive $\sigma$ annotation than using fixed $\sigma$.
These results demonstrate that taking the association between the overlap of the receptive fields and the perspective information into account has a better effect when the ground truth variation is larger.

\begin{table}[tp]
    \centering
    \caption{The performance improvements of our DRF compared with the baseline using different ground truth annotations on ShanghaiTech Part\_A.}
    \begin{tabular}{c|C{1cm}C{1cm}|C{1cm}C{1cm}}
    \hline
        \multirow{2}{*}{} &\multicolumn{2}{c|}{$\sigma=15$}&\multicolumn{2}{c}{Adaptive $\sigma$}\\
        & MAE  & RMSE  & MAE  & RMSE \\
        \hline
        Baseline & 58.1 & 90.8& 57.4 & 88.9\\
        DRF & 49.5 & 78.3& 47.1 & 74.3 \\
        \hline
        Improvement & 14.8\% & 13.8\% & 17.9\% & 16.4\%\\
        \hline
    \end{tabular}
    
    \label{tab:annotation}
\end{table}

\begin{table}[tp]
    \centering
    \caption{The performance of different ground truth annotations.}
    \begin{tabular}{c|C{1cm}C{1cm}|C{1cm}C{1cm}}
    \hline
        \multirow{2}{*}{} &\multicolumn{2}{c|}{ShanghaiTech A}&\multicolumn{2}{c}{ShanghaiTech B}\\
        & MAE  & RMSE  & MAE  & RMSE \\
        \hline
        Fixed $\sigma=15$ & 58.1 & 90.8 & 7.6  & 12.1 \\
        Adaptive $\sigma$ & 57.4 & 88.9 & 7.9 & 12.5 \\
        SDS & \textbf{53.4} & \textbf{85.5} & \textbf{7.0} & \textbf{11.0}\\
        \hline
    \end{tabular}
    
    \label{tab:SDS}
\end{table}

We use different methods of generating the ground truth annotations including our SDS and compare their performance. We present the results in Tab.~\ref{tab:SDS}. SDS achieves the best estimation results on datasets with different scales of the perspective variations due to the utilization of self-distillation and the rich perspective information integrated in the density maps.

\begin{table*}[tp]
    \centering
    \caption{Performance comparisons with state-of-the-art methods on four datatsets.}
    \begin{tabular}{c|C{1.2cm}C{1.2cm}|C{1.2cm}C{1.2cm}|C{1.2cm}C{1.2cm}|C{1.2cm}C{1.2cm}}
        \hline
        \multirow{2}{*}{Methods} &\multicolumn{2}{c|}{ShanghaiTechA} &\multicolumn{2}{c|}{ShanghaiTechB}&\multicolumn{2}{c|}{UCF\_QNRF}&\multicolumn{2}{c}{UCF\_CC\_50}\\ 
        & MAE  & RMSE  &MAE &RMSE &MAE &RMSE &MAE &RMSE \\
        \hline
        MCNN~\cite{conf/cvpr/ZhangZCGM16} & 110.2 & 173.2 &26.4 & 41.3& 277 & 426 &377.6 & 509.1\\
        Switching-CNN~\cite{conf/cvpr/SamSB17} & 90.4 & 135.0 &21.6 & 33.4& 228 & 445 &318.1 & 439.2\\
        ACSCP~\cite{conf/cvpr/ShenXNWHY18} & 75.7 & 102.7 &17.2 & 27.4&-&-&291.0 & 404.6\\
        CP-CNN~\cite{conf/iccv/SindagiP17} & 73.6 & 106.4 &20.1 & 30.1&-&-&298.8 & 320.9\\
        IG-CNN~\cite{conf/cvpr/SamSBS18} & 72.5 & 118.2 &13.6 & 21.1&-&-&291.4 & 349.4\\
        CSRNet~\cite{conf/cvpr/LiZC18} & 68.2 & 115.0 &10.6 & 16.0&-&-&266.1 & 397.5\\
        SANet~\cite{conf/eccv/CaoWZS18} & 67.0 & 104.5 &8.4 & 13.6&-&-&258.4 & 334.9\\
        CFF~\cite{conf/iccv/ShiMS19} & 65.2 & 109.4 &7.2 & 12.2& 93.8 & 146.5 &-&-\\
        SFCN~\cite{conf/cvpr/WangGL019} & 64.8 & 107.5 &7.6 & 13.0&102.0 & 171.4&214.2 & 318.2\\
        TEDNet~\cite{conf/cvpr/JiangXZZ0D019} & 64.2 & 109.1 &8.2 & 12.8&113 & 188&249.4 & 354.5\\
        ADCrowdNet~\cite{conf/cvpr/LiuLZNPW19} & 63.2 & 98.9 &7.6 & 12.9&-&-&257.1 & 363.5\\
        BL~\cite{conf/iccv/MaWHG19} & 62.8 & 101.8 &7.7 & 12.7&88.7 & 154.8&229.3 & 308.2\\
        PACNN~\cite{conf/cvpr/ShiY0C19} & 62.4 & 102.0 &7.6 & 11.8&-&-&241.7 & 320.7\\
        CAN~\cite{conf/cvpr/LiuSF19} & 62.3 & 100.0 &7.8 & 12.2&107 & 183&212.2 & 243.7\\
        RPNet~\cite{conf/cvpr/YangLWSHS20} & 61.2 & 96.9 & - & - &-&-&-&-\\
        DSSINet~\cite{conf/iccv/LiuQLLOL19} & 60.6 & 96.0 &6.9 & 10.3&99.1 & 159.2&216.9 & 302.4\\
        DM-Count~\cite{journals/corr/abs-2009-13077} & 59.7 & 95.7 & 7.4 & 11.8 & 85.6 & 148.3 & 211.0 & 291.5\\
        RANet~\cite{conf/iccv/ZhangSXZZ0019} & 59.4 & 102.0 &7.9 & 12.9&111 & 190&239.8 & 319.4\\
        SPANet~\cite{conf/iccv/ChengLD0H19}+SANet~\cite{conf/eccv/CaoWZS18} & 59.4 & 92.5 &6.5 & 9.9&-&-&232.6 & 311.7\\
        M-SFANet+M-SegNet~\cite{journals/corr/abs-2003-05586} & 57.6 & 94.5 & 6.3 & 10.1 & 85.6 & 147.8 & 162.3 & 256.3\\
        PGCNet~\cite{conf/iccv/YanYZTWWD19} & 57.0 & 86.0 &8.8 & 13.7&-&-& 244.6 & 317.6\\
        ADSCNet~\cite{conf/cvpr/BaiHQHWY20} & 55.4 & 97.7 &6.4 & 11.3&71.3 & 132.5&198.4 & 267.3\\
        \hline
        PANet & \textbf{45.2} & \textbf{73.1} & \textbf{5.9} & \textbf{9.3} & \textbf{49.1} & \textbf{106.0} & \textbf{160.3} & \textbf{223.7}\\
        \hline
    \end{tabular}
    
    \label{tab:sota}
\end{table*}

Furthermore, we implement ablation experiments on all four datasets. We carry experiments on self-distilling supervision (SDS) and dynamic receptive fields (DRF) separately. The results are illustrated in Tab.~\ref{tab:ablation}. Following ADSCNet~\cite{conf/cvpr/BaiHQHWY20}, our baseline is CSRNet*, which is CSRNet~\cite{conf/cvpr/LiZC18} with Batch Normalization layers and outperforms the original CSRNet. The network is trained with Adam optimizer instead of SGD optimizer, which further increases the accuracy, following some papers like BL~\cite{conf/iccv/MaWHG19}. We use L1 loss instead of L2 loss, which is the same as ADSCNet~\cite{conf/cvpr/BaiHQHWY20}. This loss function also improves the performance of the baseline. In the experiments on SDS exclusively, the refined dilated convolutional layers in the precise network are replaced with vanilla dilated convolutional layers, where the dilation rate is a constant set as $2$. When we validate the effectiveness of DRF alone, we use Gaussian annotated density maps as the ground truth and the refined density maps serve as the final estimation results.

From the ablation study results, it is demonstrated that both our SDS and DRF frameworks can improve MAE and RMSE results on all four datasets.
Combining the two proposed approaches, our PANet achieves better performance than each separate framework. On ShanghaiTech dataset, PANet decreases MAE by 21.3\% and RMSE by 17.8\% on Part A and decreases MAE by 22.4\% and RMSE by 23.1\% on Part B, compared with the baseline. There are more significant improvements on the other two datasets. PANet improves the baseline with relative MAE and RMSE improvements of 57.6\% and 42.8\% on the UCF\_QNRF dataset, 35.9\% and 33.3\% on the UCF\_CC\_50 dataset. Some visualization examples on ShanghaiTech Part\_A are shown in Fig.~\ref{fig:perform}.

\subsection{Comparisons with State-of-the-art}

Tab.~\ref{tab:sota} reports the results of PANet compared with some state-of-the-art methods on ShanghaiTech Part\_A, ShanghaiTech Part\_B, UCF\_QNRF, and UCF\_CC\_50 datasets. Our method makes a breakthrough in the performance on these four datasets, considering both MAE and RMSE. Compared with the state-of-the-art methods, PANet improves the MAE and RMSE performance by 31.1\% and 20.0\% on UCF\_QNRF, 18.4\% and 15.0\% on ShanghaiTech Part\_A, 6.3\% and 6.1\% on ShanghaiTech Part\_B, 1.2\% and 8.2\% on UCF\_CC\_50. The considerable progress on these datasets confirms the generalization capability of our PANet.

\section{Conclusion}

In this paper, we present a novel perspective-aware learning framework named PANet for crowd counting, which consists of two major components. Based on the observation that smaller receptive fields ought to be used in far-end regions and vice versa, the dynamic receptive fields (DRF) framework learns the scales of receptive fields by supervision from rough density maps. DRF is more explainable and outperforms the existing methods of adjusting receptive fields.
Motivated by knowledge distillation, the proposed two-stage framework, self-distilling supervision (SDS), refines the ground truth density map guided by the prediction in the first stage. 
The refined density map integrates rich perspective information and improves the precise network in the second stage.

Comprehensive experimental results support our assumptions and validate the effectiveness of each component of our method. Combining SDS and DRF, our PANet achieves state-of-the-art performance in terms of both MAE and RMSE on four challenging crowd counting datasets.

{\small
\bibliographystyle{ieee_fullname}
\bibliography{paper}

\begin{thebibliography}{10}\itemsep=-1pt

\bibitem{conf/cvpr/BaiHQHWY20}
Shuai Bai, Zhiqun He, Yu Qiao, Hanzhe Hu, Wei Wu, and Junjie Yan.
\newblock Adaptive dilated network with self-correction supervision for
  counting.
\newblock In {\em 2020 {IEEE/CVF} Conference on Computer Vision and Pattern
  Recognition, {CVPR} 2020, Seattle, WA, USA, June 13-19, 2020}, pages
  4593--4602. {IEEE}, 2020.

\bibitem{conf/mm/BoominathanKB16}
Lokesh Boominathan, Srinivas S.~S. Kruthiventi, and R.~Venkatesh Babu.
\newblock Crowdnet: A deep convolutional network for dense crowd counting.
\newblock In Alan Hanjalic, Cees Snoek, Marcel Worring, Dick C.~A. Bulterman,
  Benoit Huet, Aisling Kelliher, Yiannis Kompatsiaris, and Jin Li, editors,
  {\em ACM Multimedia}, pages 640--644. ACM, 2016.

\bibitem{conf/cvpr/BrostowC06}
Gabriel~J. Brostow and Roberto Cipolla.
\newblock Unsupervised bayesian detection of independent motion in crowds.
\newblock In {\em CVPR (1)}, pages 594--601. IEEE Computer Society, 2006.

\bibitem{conf/kdd/BucilaCN06}
Cristian Bucila, Rich Caruana, and Alexandru Niculescu-Mizil.
\newblock Model compression.
\newblock In Tina Eliassi-Rad, Lyle~H. Ungar, Mark Craven, and Dimitrios
  Gunopulos, editors, {\em KDD}, pages 535--541. ACM, 2006.

\bibitem{conf/eccv/CaoWZS18}
Xinkun Cao, Zhipeng Wang, Yanyun Zhao, and Fei Su.
\newblock Scale aggregation network for accurate and efficient crowd counting.
\newblock In Vittorio Ferrari, Martial Hebert, Cristian Sminchisescu, and Yair
  Weiss, editors, {\em ECCV (5)}, volume 11209 of {\em Lecture Notes in
  Computer Science}, pages 757--773. Springer, 2018.

\bibitem{conf/iccv/ChengLD0H19}
Zhi{-}Qi Cheng, Jun{-}Xiu Li, Qi Dai, Xiao Wu, and Alexander~G. Hauptmann.
\newblock Learning spatial awareness to improve crowd counting.
\newblock In {\em 2019 {IEEE/CVF} International Conference on Computer Vision,
  {ICCV} 2019, Seoul, Korea (South), October 27 - November 2, 2019}, pages
  6151--6160. {IEEE}, 2019.

\bibitem{conf/mm/ChengLD0HH19}
Zhi-Qi Cheng, Jun-Xiu Li, Qi Dai, Xiao Wu, Jun-Yan He, and Alexander~G.
  Hauptmann.
\newblock Improving the learning of multi-column convolutional neural network
  for crowd counting.
\newblock In Laurent Amsaleg, Benoit Huet, Martha~A. Larson, Guillaume Gravier,
  Hayley Hung, Chong-Wah Ngo, and Wei~Tsang Ooi, editors, {\em ACM Multimedia},
  pages 1897--1906. ACM, 2019.

\bibitem{conf/iccv/DaiQXLZHW17}
Jifeng Dai, Haozhi Qi, Yuwen Xiong, Yi Li, Guodong Zhang, Han Hu, and Yichen
  Wei.
\newblock Deformable convolutional networks.
\newblock In {\em {IEEE} International Conference on Computer Vision, {ICCV}
  2017, Venice, Italy, October 22-29, 2017}, pages 764--773. {IEEE} Computer
  Society, 2017.

\bibitem{conf/eccv/DollarBBPT08}
Piotr Dollár, Boris Babenko, Serge~J. Belongie, Pietro Perona, and Zhuowen Tu.
\newblock Multiple component learning for object detection.
\newblock In David~A. Forsyth, Philip H.~S. Torr, and Andrew Zisserman,
  editors, {\em ECCV (2)}, volume 5303 of {\em Lecture Notes in Computer
  Science}, pages 211--224. Springer, 2008.

\bibitem{journals/pami/EnzweilerG09}
Markus Enzweiler and Dariu~M. Gavrila.
\newblock Monocular pedestrian detection: Survey and experiments.
\newblock {\em IEEE Trans. Pattern Anal. Mach. Intell.}, 31(12):2179--2195,
  2009.

\bibitem{journals/eaai/FuXLLYZ15}
Min Fu, Pei Xu, Xudong Li, Qihe Liu, Mao Ye, and Ce Zhu.
\newblock Fast crowd density estimation with convolutional neural networks.
\newblock {\em Eng. Appl. Artif. Intell.}, 43:81--88, 2015.

\bibitem{conf/cvpr/GeC09}
Weina Ge and Robert~T. Collins.
\newblock Marked point processes for crowd counting.
\newblock In {\em 2009 {IEEE} Computer Society Conference on Computer Vision
  and Pattern Recognition {(CVPR} 2009), 20-25 June 2009, Miami, Florida,
  {USA}}, pages 2913--2920. {IEEE} Computer Society, 2009.

\bibitem{conf/mm/GuoLZW19}
Dan Guo, Kun Li, Zheng-Jun Zha, and Meng Wang.
\newblock Dadnet: Dilated-attention-deformable convnet for crowd counting.
\newblock In Laurent Amsaleg, Benoit Huet, Martha~A. Larson, Guillaume Gravier,
  Hayley Hung, Chong-Wah Ngo, and Wei~Tsang Ooi, editors, {\em ACM Multimedia},
  pages 1823--1832. ACM, 2019.

\bibitem{hinton2015distilling}
Geoffrey Hinton, Oriol Vinyals, and Jeff Dean.
\newblock Distilling the knowledge in a neural network, 2015.
\newblock cite arxiv:1503.02531Comment: NIPS 2014 Deep Learning Workshop.

\bibitem{conf/cvpr/IdreesSSS13}
Haroon Idrees, Imran Saleemi, Cody Seibert, and Mubarak Shah.
\newblock Multi-source multi-scale counting in extremely dense crowd images.
\newblock In {\em 2013 {IEEE} Conference on Computer Vision and Pattern
  Recognition, Portland, OR, USA, June 23-28, 2013}, pages 2547--2554. {IEEE}
  Computer Society, 2013.

\bibitem{conf/eccv/IdreesTAZARS18}
Haroon Idrees, Muhmmad Tayyab, Kishan Athrey, Dong Zhang, Somaya Al-Máadeed,
  Nasir~M. Rajpoot, and Mubarak Shah.
\newblock Composition loss for counting, density map estimation and
  localization in dense crowds.
\newblock In Vittorio Ferrari, Martial Hebert, Cristian Sminchisescu, and Yair
  Weiss, editors, {\em ECCV (2)}, volume 11206 of {\em Lecture Notes in
  Computer Science}, pages 544--559. Springer, 2018.

\bibitem{conf/icml/IoffeS15}
Sergey Ioffe and Christian Szegedy.
\newblock Batch normalization: Accelerating deep network training by reducing
  internal covariate shift.
\newblock In Francis~R. Bach and David~M. Blei, editors, {\em Proceedings of
  the 32nd International Conference on Machine Learning, {ICML} 2015, Lille,
  France, 6-11 July 2015}, volume~37 of {\em {JMLR} Workshop and Conference
  Proceedings}, pages 448--456. JMLR.org, 2015.

\bibitem{conf/cvpr/JiangXZZ0D019}
Xiaolong Jiang, Zehao Xiao, Baochang Zhang, Xiantong Zhen, Xianbin Cao,
  David~S. Doermann, and Ling Shao.
\newblock Crowd counting and density estimation by trellis encoder-decoder
  networks.
\newblock In {\em {IEEE} Conference on Computer Vision and Pattern Recognition,
  {CVPR} 2019, Long Beach, CA, USA, June 16-20, 2019}, pages 6133--6142.
  Computer Vision Foundation / {IEEE}, 2019.

\bibitem{conf/cvpr/JiangZXZLZYP20}
Xiaoheng Jiang, Li Zhang, Mingliang Xu, Tianzhu Zhang, Pei Lv, Bing Zhou, Xin
  Yang, and Yanwei Pang.
\newblock Attention scaling for crowd counting.
\newblock In {\em 2020 {IEEE/CVF} Conference on Computer Vision and Pattern
  Recognition, {CVPR} 2020, Seattle, WA, USA, June 13-19, 2020}, pages
  4705--4714. {IEEE}, 2020.

\bibitem{journals/corr/KingmaB14}
Diederik~P. Kingma and Jimmy Ba.
\newblock Adam: {A} method for stochastic optimization.
\newblock In Yoshua Bengio and Yann LeCun, editors, {\em 3rd International
  Conference on Learning Representations, {ICLR} 2015, San Diego, CA, USA, May
  7-9, 2015, Conference Track Proceedings}, 2015.

\bibitem{conf/cvpr/LeibeSS05}
Bastian Leibe, Edgar Seemann, and Bernt Schiele.
\newblock Pedestrian detection in crowded scenes.
\newblock In {\em CVPR (1)}, pages 878--885. IEEE Computer Society, 2005.

\bibitem{conf/icpr/LiZHT08}
Min Li, Zhaoxiang Zhang, Kaiqi Huang, and Tieniu Tan.
\newblock Estimating the number of people in crowded scenes by mid based
  foreground segmentation and head-shoulder detection.
\newblock In {\em ICPR}, pages 1--4. IEEE Computer Society, 2008.

\bibitem{conf/cvpr/LiZC18}
Yuhong Li, Xiaofan Zhang, and Deming Chen.
\newblock Csrnet: Dilated convolutional neural networks for understanding the
  highly congested scenes.
\newblock In {\em 2018 {IEEE} Conference on Computer Vision and Pattern
  Recognition, {CVPR} 2018, Salt Lake City, UT, USA, June 18-22, 2018}, pages
  1091--1100. {IEEE} Computer Society, 2018.

\bibitem{journals/tsmc/LinCC01}
Sheng-Fuu Lin, Jaw-Yeh Chen, and Hung-Xin Chao.
\newblock Estimation of number of people in crowded scenes using perspective
  transformation.
\newblock {\em IEEE Trans. Syst. Man Cybern. Part A}, 31(6):645--654, 2001.

\bibitem{conf/cvpr/LiuWM19}
Chenchen Liu, Xinyu Weng, and Yadong Mu.
\newblock Recurrent attentive zooming for joint crowd counting and precise
  localization.
\newblock In {\em {IEEE} Conference on Computer Vision and Pattern Recognition,
  {CVPR} 2019, Long Beach, CA, USA, June 16-20, 2019}, pages 1217--1226.
  Computer Vision Foundation / {IEEE}, 2019.

\bibitem{conf/cvpr/0011GMH18}
Jiang Liu, Chenqiang Gao, Deyu Meng, and Alexander~G. Hauptmann.
\newblock Decidenet: Counting varying density crowds through attention guided
  detection and density estimation.
\newblock In {\em 2018 {IEEE} Conference on Computer Vision and Pattern
  Recognition, {CVPR} 2018, Salt Lake City, UT, USA, June 18-22, 2018}, pages
  5197--5206. {IEEE} Computer Society, 2018.

\bibitem{conf/iccv/LiuQLLOL19}
Lingbo Liu, Zhilin Qiu, Guanbin Li, Shufan Liu, Wanli Ouyang, and Liang Lin.
\newblock Crowd counting with deep structured scale integration network.
\newblock In {\em 2019 {IEEE/CVF} International Conference on Computer Vision,
  {ICCV} 2019, Seoul, Korea (South), October 27 - November 2, 2019}, pages
  1774--1783. {IEEE}, 2019.

\bibitem{conf/ijcai/LiuWLOL18}
Lingbo Liu, Hongjun Wang, Guanbin Li, Wanli Ouyang, and Liang Lin.
\newblock Crowd counting using deep recurrent spatial-aware network.
\newblock In J{\'{e}}r{\^{o}}me Lang, editor, {\em Proceedings of the
  Twenty-Seventh International Joint Conference on Artificial Intelligence,
  {IJCAI} 2018, July 13-19, 2018, Stockholm, Sweden}, pages 849--855.
  ijcai.org, 2018.

\bibitem{conf/cvpr/LiuLZNPW19}
Ning Liu, Yongchao Long, Changqing Zou, Qun Niu, Li Pan, and Hefeng Wu.
\newblock Adcrowdnet: An attention-injective deformable convolutional network
  for crowd understanding.
\newblock In {\em {IEEE} Conference on Computer Vision and Pattern Recognition,
  {CVPR} 2019, Long Beach, CA, USA, June 16-20, 2019}, pages 3225--3234.
  Computer Vision Foundation / {IEEE}, 2019.

\bibitem{conf/cvpr/LiuSF19}
Weizhe Liu, Mathieu Salzmann, and Pascal Fua.
\newblock Context-aware crowd counting.
\newblock In {\em {IEEE} Conference on Computer Vision and Pattern Recognition,
  {CVPR} 2019, Long Beach, CA, USA, June 16-20, 2019}, pages 5099--5108.
  Computer Vision Foundation / {IEEE}, 2019.

\bibitem{conf/cvpr/LiuSZW19}
Yuting Liu, Miaojing Shi, Qijun Zhao, and Xiaofang Wang.
\newblock Point in, box out: Beyond counting persons in crowds.
\newblock In {\em {IEEE} Conference on Computer Vision and Pattern Recognition,
  {CVPR} 2019, Long Beach, CA, USA, June 16-20, 2019}, pages 6469--6478.
  Computer Vision Foundation / {IEEE}, 2019.

\bibitem{conf/iccv/MaWHG19}
Zhiheng Ma, Xing Wei, Xiaopeng Hong, and Yihong Gong.
\newblock Bayesian loss for crowd count estimation with point supervision.
\newblock In {\em 2019 {IEEE/CVF} International Conference on Computer Vision,
  {ICCV} 2019, Seoul, Korea (South), October 27 - November 2, 2019}, pages
  6141--6150. {IEEE}, 2019.

\bibitem{conf/eccv/Onoro-RubioL16}
Daniel Oñoro-Rubio and Roberto~Javier López-Sastre.
\newblock Towards perspective-free object counting with deep learning.
\newblock In Bastian Leibe, Jiri Matas, Nicu Sebe, and Max Welling, editors,
  {\em ECCV (7)}, volume 9911 of {\em Lecture Notes in Computer Science}, pages
  615--629. Springer, 2016.

\bibitem{journals/ijcv/RussakovskyDSKS15}
Olga Russakovsky, Jia Deng, Hao Su, Jonathan Krause, Sanjeev Satheesh, Sean Ma,
  Zhiheng Huang, Andrej Karpathy, Aditya Khosla, Michael~S. Bernstein,
  Alexander~C. Berg, and Fei-Fei Li.
\newblock Imagenet large scale visual recognition challenge.
\newblock {\em Int. J. Comput. Vis.}, 115(3):211--252, 2015.

\bibitem{conf/aaai/SamB18}
Deepak~Babu Sam and R.~Venkatesh Babu.
\newblock Top-down feedback for crowd counting convolutional neural network.
\newblock In Sheila~A. McIlraith and Kilian~Q. Weinberger, editors, {\em
  Proceedings of the Thirty-Second {AAAI} Conference on Artificial
  Intelligence, (AAAI-18), the 30th innovative Applications of Artificial
  Intelligence (IAAI-18), and the 8th {AAAI} Symposium on Educational Advances
  in Artificial Intelligence (EAAI-18), New Orleans, Louisiana, USA, February
  2-7, 2018}, pages 7323--7330. {AAAI} Press, 2018.

\bibitem{conf/cvpr/SamSBS18}
Deepak~Babu Sam, Neeraj~N. Sajjan, R.~Venkatesh Babu, and Mukundhan Srinivasan.
\newblock Divide and grow: Capturing huge diversity in crowd images with
  incrementally growing {CNN}.
\newblock In {\em 2018 {IEEE} Conference on Computer Vision and Pattern
  Recognition, {CVPR} 2018, Salt Lake City, UT, USA, June 18-22, 2018}, pages
  3618--3626. {IEEE} Computer Society, 2018.

\bibitem{conf/cvpr/SamSB17}
Deepak~Babu Sam, Shiv Surya, and R.~Venkatesh Babu.
\newblock Switching convolutional neural network for crowd counting.
\newblock In {\em 2017 {IEEE} Conference on Computer Vision and Pattern
  Recognition, {CVPR} 2017, Honolulu, HI, USA, July 21-26, 2017}, pages
  4031--4039. {IEEE} Computer Society, 2017.

\bibitem{conf/cvpr/ShenXNWHY18}
Zan Shen, Yi Xu, Bingbing Ni, Minsi Wang, Jianguo Hu, and Xiaokang Yang.
\newblock Crowd counting via adversarial cross-scale consistency pursuit.
\newblock In {\em 2018 {IEEE} Conference on Computer Vision and Pattern
  Recognition, {CVPR} 2018, Salt Lake City, UT, USA, June 18-22, 2018}, pages
  5245--5254. {IEEE} Computer Society, 2018.

\bibitem{conf/cvpr/ShiY0C19}
Miaojing Shi, Zhaohui Yang, Chao Xu, and Qijun Chen.
\newblock Revisiting perspective information for efficient crowd counting.
\newblock In {\em {IEEE} Conference on Computer Vision and Pattern Recognition,
  {CVPR} 2019, Long Beach, CA, USA, June 16-20, 2019}, pages 7279--7288.
  Computer Vision Foundation / {IEEE}, 2019.

\bibitem{conf/iccv/ShiMS19}
Zenglin Shi, Pascal Mettes, and Cees Snoek.
\newblock Counting with focus for free.
\newblock In {\em 2019 {IEEE/CVF} International Conference on Computer Vision,
  {ICCV} 2019, Seoul, Korea (South), October 27 - November 2, 2019}, pages
  4199--4208. {IEEE}, 2019.

\bibitem{conf/cvpr/ShiZLCYCZ18}
Zenglin Shi, Le Zhang, Yun Liu, Xiaofeng Cao, Yangdong Ye, Ming{-}Ming Cheng,
  and Guoyan Zheng.
\newblock Crowd counting with deep negative correlation learning.
\newblock In {\em 2018 {IEEE} Conference on Computer Vision and Pattern
  Recognition, {CVPR} 2018, Salt Lake City, UT, USA, June 18-22, 2018}, pages
  5382--5390. {IEEE} Computer Society, 2018.

\bibitem{journals/corr/SimonyanZ14a}
Karen Simonyan and Andrew Zisserman.
\newblock Very deep convolutional networks for large-scale image recognition.
\newblock In Yoshua Bengio and Yann LeCun, editors, {\em 3rd International
  Conference on Learning Representations, {ICLR} 2015, San Diego, CA, USA, May
  7-9, 2015, Conference Track Proceedings}, 2015.

\bibitem{conf/iccv/SindagiP17}
Vishwanath~A. Sindagi and Vishal~M. Patel.
\newblock Generating high-quality crowd density maps using contextual pyramid
  cnns.
\newblock In {\em {IEEE} International Conference on Computer Vision, {ICCV}
  2017, Venice, Italy, October 22-29, 2017}, pages 1879--1888. {IEEE} Computer
  Society, 2017.

\bibitem{journals/corr/abs-2003-05586}
Pongpisit Thanasutives, Ken ichi Fukui, Masayuki Numao, and Boonserm
  Kijsirikul.
\newblock Encoder-decoder based convolutional neural networks with
  multi-scale-aware modules for crowd counting.
\newblock {\em CoRR}, abs/2003.05586, 2020.

\bibitem{conf/avss/TopkayaEP14}
Ibrahim~Saygin Topkaya, Hakan Erdogan, and Fatih~Murat Porikli.
\newblock Counting people by clustering person detector outputs.
\newblock In {\em AVSS}, pages 313--318. IEEE Computer Society, 2014.

\bibitem{conf/eccv/WalachW16}
Elad Walach and Lior Wolf.
\newblock Learning to count with cnn boosting.
\newblock In Bastian Leibe, Jiri Matas, Nicu Sebe, and Max Welling, editors,
  {\em ECCV (2)}, volume 9906 of {\em Lecture Notes in Computer Science}, pages
  660--676. Springer, 2016.

\bibitem{wan2020modeling}
Jia Wan and Antoni Chan.
\newblock Modeling noisy annotations for crowd counting.
\newblock {\em Advances in Neural Information Processing Systems}, 33, 2020.

\bibitem{journals/corr/abs-2009-13077}
Boyu Wang, Huidong Liu, Dimitris Samaras, and Minh~Hoai Nguyen.
\newblock Distribution matching for crowd counting.
\newblock In Hugo Larochelle, Marc'Aurelio Ranzato, Raia Hadsell,
  Maria{-}Florina Balcan, and Hsuan{-}Tien Lin, editors, {\em Advances in
  Neural Information Processing Systems 33: Annual Conference on Neural
  Information Processing Systems 2020, NeurIPS 2020, December 6-12, 2020,
  virtual}, 2020.

\bibitem{conf/mm/WangZYLC15}
Chuan Wang, Hua Zhang, Liang Yang, Si Liu, and Xiaochun Cao.
\newblock Deep people counting in extremely dense crowds.
\newblock In Xiaofang Zhou, Alan~F. Smeaton, Qi Tian, Dick C.~A. Bulterman,
  Heng~Tao Shen, Ketan Mayer-Patel, and Shuicheng Yan, editors, {\em ACM
  Multimedia}, pages 1299--1302. ACM, 2015.

\bibitem{conf/cvpr/WangGL019}
Qi Wang, Junyu Gao, Wei Lin, and Yuan Yuan.
\newblock Learning from synthetic data for crowd counting in the wild.
\newblock In {\em {IEEE} Conference on Computer Vision and Pattern Recognition,
  {CVPR} 2019, Long Beach, CA, USA, June 16-20, 2019}, pages 8198--8207.
  Computer Vision Foundation / {IEEE}, 2019.

\bibitem{conf/iccv/XiongLLLCS19}
Haipeng Xiong, Hao Lu, Chengxin Liu, Liang Liu, Zhiguo Cao, and Chunhua Shen.
\newblock From open set to closed set: Counting objects by spatial
  divide-and-conquer.
\newblock In {\em 2019 {IEEE/CVF} International Conference on Computer Vision,
  {ICCV} 2019, Seoul, Korea (South), October 27 - November 2, 2019}, pages
  8361--8370. {IEEE}, 2019.

\bibitem{conf/iccv/YanYZTWWD19}
Zhaoyi Yan, Yuchen Yuan, Wangmeng Zuo, Xiao Tan, Yezhen Wang, Shilei Wen, and
  Errui Ding.
\newblock Perspective-guided convolution networks for crowd counting.
\newblock In {\em 2019 {IEEE/CVF} International Conference on Computer Vision,
  {ICCV} 2019, Seoul, Korea (South), October 27 - November 2, 2019}, pages
  952--961. {IEEE}, 2019.

\bibitem{conf/cvpr/YangLWSHS20}
Yifan Yang, Guorong Li, Zhe Wu, Li Su, Qingming Huang, and Nicu Sebe.
\newblock Reverse perspective network for perspective-aware object counting.
\newblock In {\em 2020 {IEEE/CVF} Conference on Computer Vision and Pattern
  Recognition, {CVPR} 2020, Seattle, WA, USA, June 13-19, 2020}, pages
  4373--4382. {IEEE}, 2020.

\bibitem{conf/iccv/ZhangSXZZ0019}
Anran Zhang, Jiayi Shen, Zehao Xiao, Fan Zhu, Xiantong Zhen, Xianbin Cao, and
  Ling Shao.
\newblock Relational attention network for crowd counting.
\newblock In {\em 2019 {IEEE/CVF} International Conference on Computer Vision,
  {ICCV} 2019, Seoul, Korea (South), October 27 - November 2, 2019}, pages
  6787--6796. {IEEE}, 2019.

\bibitem{conf/cvpr/ZhangZCGM16}
Yingying Zhang, Desen Zhou, Siqin Chen, Shenghua Gao, and Yi Ma.
\newblock Single-image crowd counting via multi-column convolutional neural
  network.
\newblock In {\em 2016 {IEEE} Conference on Computer Vision and Pattern
  Recognition, {CVPR} 2016, Las Vegas, NV, USA, June 27-30, 2016}, pages
  589--597. {IEEE} Computer Society, 2016.

\bibitem{conf/cvpr/ZhaoN03}
Tao Zhao and Ramakant Nevatia.
\newblock Bayesian human segmentation in crowded situations.
\newblock In {\em CVPR (2)}, pages 459--466. IEEE Computer Society, 2003.

\end{thebibliography}
}

\end{document}